%% file: acl_latex.tex
\pdfoutput=1

\documentclass[11pt]{article}

\usepackage[preprint]{acl}

\usepackage{times}
\usepackage{latexsym}
\usepackage{paralist, tabularx}
\usepackage{multicol}
\usepackage{multirow}
\usepackage{amsmath}
\usepackage{booktabs}
\usepackage{subcaption}
\usepackage{soul}
\usepackage{color}
\usepackage{xcolor}

\usepackage{authblk}

\usepackage[T1]{fontenc}

\usepackage[utf8]{inputenc}

\usepackage{microtype}

\usepackage{inconsolata}

\usepackage{graphicx}

\title{Getting More from Less: Large Language Models are Good 

Spontaneous Multilingual Learners}

\newcommand*{\affaddr}[1]{#1} 
\newcommand*{\affmark}[1][*]{\textsuperscript{#1}}

\author{
\textbf{Shimao Zhang}\affmark[$\clubsuit$], \textbf{Changjiang Gao}\affmark[$\clubsuit$], \textbf{Wenhao Zhu}\affmark[$\clubsuit$], \textbf{Jiajun Chen}\affmark[$\clubsuit$], \textbf{Xin Huang}\affmark[$\Diamond$], \\
\textbf{Xue Han}\affmark[$\Diamond$], \textbf{Junlan Feng}\affmark[$\Diamond$], \textbf{Chao Deng}\affmark[$\Diamond$], \textbf{Shujian Huang}\affmark[$\clubsuit$]\thanks{Corresponding author} \\
\affaddr{\affmark[$\clubsuit$]National Key Laboratory for Novel Software Technology, Nanjing University} \\
\affaddr{\affmark[$\Diamond$]China Mobile Research Beijing, China} \\
\texttt{\{smzhang,gaocj,zhuwh\}@smail.nju.edu.cn, \{huangsj,chenjj\}@nju.edu.cn} \\
\texttt{\{huangxinyjy,hanxueai,fengjunlan,dengchao\}@chinamobile.com}
}

\begin{document}
\maketitle
\input{Chapters/00-abstract}

\input{Chapters/01-introduction}

\input{Chapters/02-background}

\input{Chapters/03-methods}

\input{Chapters/04-experimental-setup}

\input{Chapters/05-results}

\input{Chapters/06-further-analysis}

\input{Chapters/07-conclusion}

\bibliography{custom}

\newpage

\appendix

\section{LLaMA 2 Language Distribution}\label{appendix:llama2-lang-distribution}
\begin{table}[htbp]
    \centering
    \begin{tabular}{lc}
        \toprule
        \textbf{Language} & \textbf{Percent} \\
        \midrule
        en & 89.70\% \\
        unknown & 8.38\% \\
        de & 0.17\% \\
        fr & 0.16\% \\
        sv & 0.15\% \\
        zh & 0.13\% \\
        es & 0.13\% \\
        ru & 0.13\% \\
        nl & 0.12\% \\
        it & 0.11\% \\
        ja & 0.10\% \\
        \bottomrule
    \end{tabular}
    \caption{Top-10 (except unknown) lanaguage distribution in LLaMA-2's pretraining data~\citep{touvron2023llama}. The majority of these data is English data. And the unknown category is partially made up of programming code data.}
    \label{tab:LLaMA2-lang-percentage}
\end{table}

\section{Additional Experimental Implementations}\label{appendix:experimental-implementations}
For instruction-tuning process we mentioned above, we use LoRA (rank = 8, $\alpha$ = 16) with 3 epochs (1 epoch for PAWS to mitigate overfitting), batch\_size = 16, learning\_rate = 5e-5, val\_size = 0.05, lr\_scheduler\_type = cosine, cutoff\_len = 2048 based on the settings of LLaMA-Factory\footnote{\url{https://github.com/hiyouga/LLaMA-Factory}}~\citep{zheng2024llamafactory}, a widely used and recognized open-source project for LLMs efficient fine-tuning. We use single NVIDIA RTX A6000 48GB or single NVIDIA Tesla V100 SXM2 32GB for training. Training time varies from 4 hours to 10+ hours depending on the language and the total instance quantity.

We construct 10k parallel data for every language pair used for training. For example, for "zh/de-en" setting of Mistral-7B, we construct a dataset including 10k Chinese-to-English translation instances and a dataset including 10k German-to-English translation instances firstly. Then we instruction-tune the Mistral-7B model only on 20k randomly mixed translation data.

We use test data and few-shot examples translated from English by Google Translate for all languages to minimize the impact of test dataset and few-shot examples themselves and ensure testing fairness across different languages. We choose the few-shot examples which are not in our training data and test data. 

Additionally, we find that not only non-English inputs but also non-English outputs have significant impacts on the model's performance. For example, for Mistral-7B and emotion classification task, the accuracy on Hindi is 0.5 if we use outputs in Hindi, while the accuracy is 0.816 if the output is "positive" or "negative". This implies that generating content in the target language is another great challenge for LLMs, which is distinct from understanding and solving problems in the corresponding language. Considering we mainly focus on the language understanding and task solving capabilities, we use English outputs uniformly if it is not specified.

\section{Pearson Correlation Coefficient Based on PCA}\label{appendix:pearson-pca}

\begin{table}[htbp]
    \centering
    \begin{tabular}{l|cc}
        \toprule
        \textbf{Layer 20} & \textbf{Base} & \textbf{Trained} \\
        \midrule
        en-de & 0.9727 & 0.9752 \\
        en-fr & 0.9804 & 0.9822 \\
        en-hi & 0.9268 & 0.9526 \\
        de-fr & 0.9825 & 0.9834 \\
        de-hi & 0.9564 & 0.9707 \\
        fr-hi & 0.9518 & 0.9674 \\
        en-th & 0.8727 & 0.8941 \\
        en-sw & 0.9501 & 0.9594 \\
        en-ms & 0.9552 & 0.9620 \\
        \bottomrule
    \end{tabular}
    \caption{Pearson correlation coefficient of 1 dimension PCA results in Mistral-7B layer 20.}
    \label{tab:pearson-mistral-20}
\end{table}

\begin{table}[htbp]
    \centering
    \begin{tabular}{l|cc}
        \toprule
        \textbf{Layer 25} & \textbf{Base} & \textbf{Trained} \\
        \midrule
        en-de & 0.9286 & 0.9275 \\
        en-fr & 0.9520 & 0.9447 \\
        en-hi & 0.5060 & 0.7942 \\
        de-fr & 0.9628 & 0.9754 \\
        de-hi & 0.6233 & 0.9102 \\
        fr-hi & 0.6272 & 0.9001 \\
        en-th & 0.6791 & 0.7484 \\
        en-sw & 0.2316 & 0.8514 \\
        en-ms & 0.7835 & 0.8448 \\
        \bottomrule
    \end{tabular}
    \caption{Pearson correlation coefficient of 1 dimension PCA results in Mistral-7B layer 25.}
    \label{tab:pearson-mistral-25}
\end{table}

\section{Logit Lens and PCA Results for Qwen1.5}\label{appendix:pca-pearson-qwen}
\begin{figure*}[tbp]
    \centering
    \begin{minipage}[b]{0.32\textwidth}
    \includegraphics[width=1.0\linewidth]{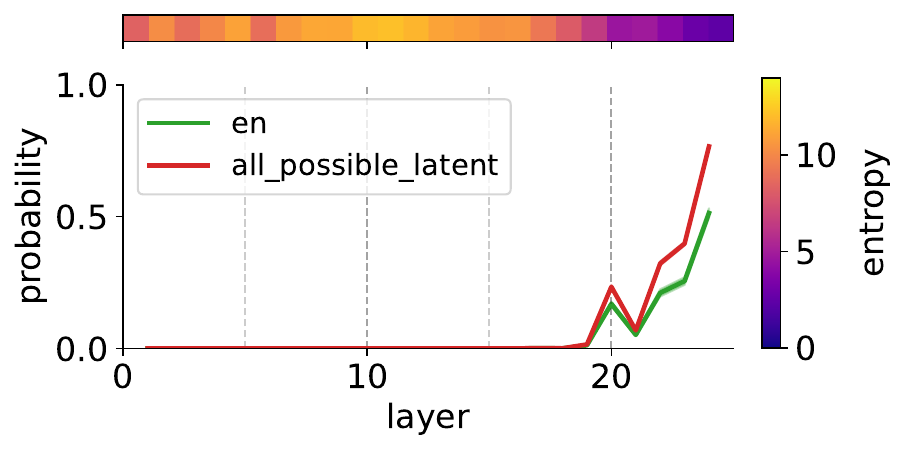}
    \subcaption{English Before}
    \label{fig:qwen1.8b-en-before}
    \end{minipage}
    \begin{minipage}[b]{0.32\textwidth}
    \includegraphics[width=1.0\linewidth]{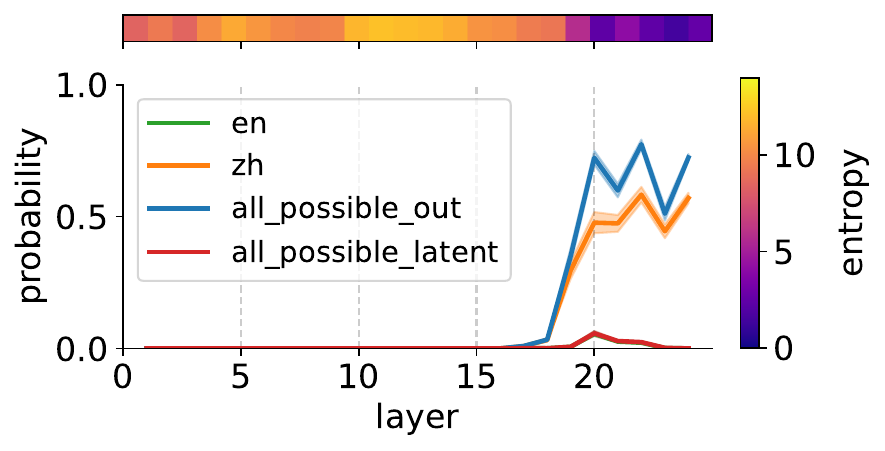}
    \subcaption{Chinese Before}
    \label{fig:qwen1.8b-zh-before}
    \end{minipage}
    \begin{minipage}[b]{0.32\textwidth}
    \includegraphics[width=1.0\linewidth]{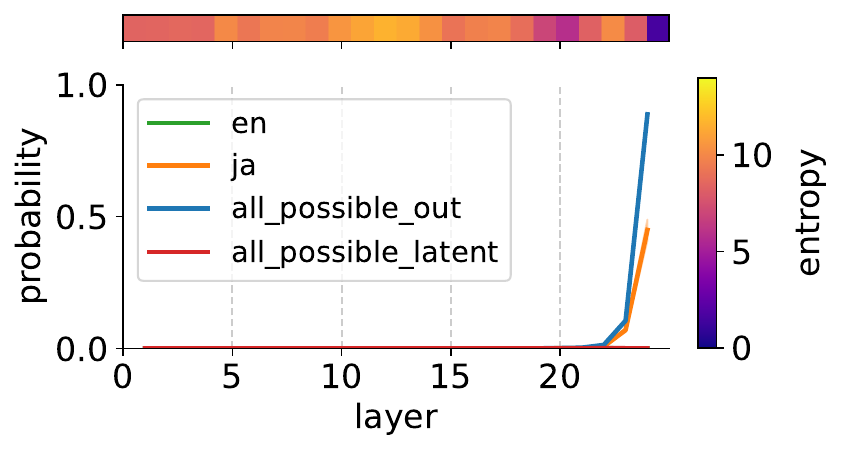}
    \subcaption{Japanese Before}
    \label{fig:qwen1.8b-ja-before}
    \end{minipage}
    \begin{minipage}[b]{0.32\textwidth}
    \includegraphics[width=1.0\linewidth]{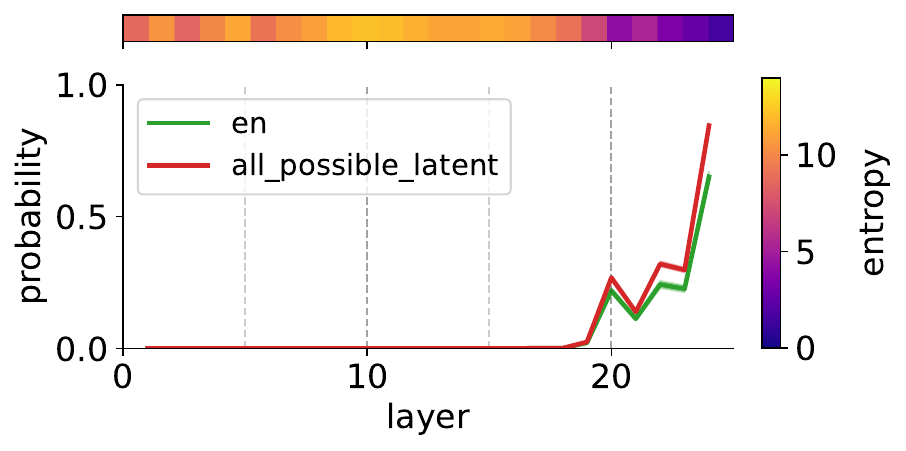}
    \subcaption{English After}
    \label{fig:qwen1.8b-en-after}
    \end{minipage}
    \begin{minipage}[b]{0.32\textwidth}
    \includegraphics[width=1.0\linewidth]{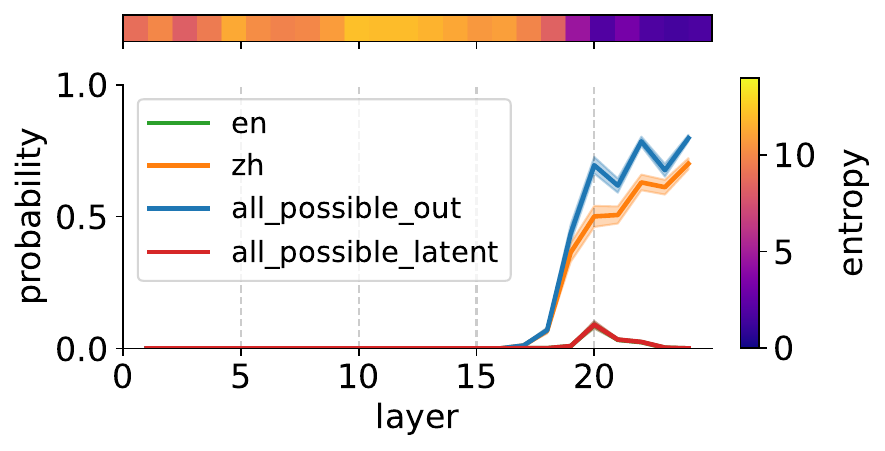}
    \subcaption{Chinese After}
    \label{fig:qwen1.8b-zh-after}
    \end{minipage}
    \begin{minipage}[b]{0.32\textwidth}
    \includegraphics[width=1.0\linewidth]{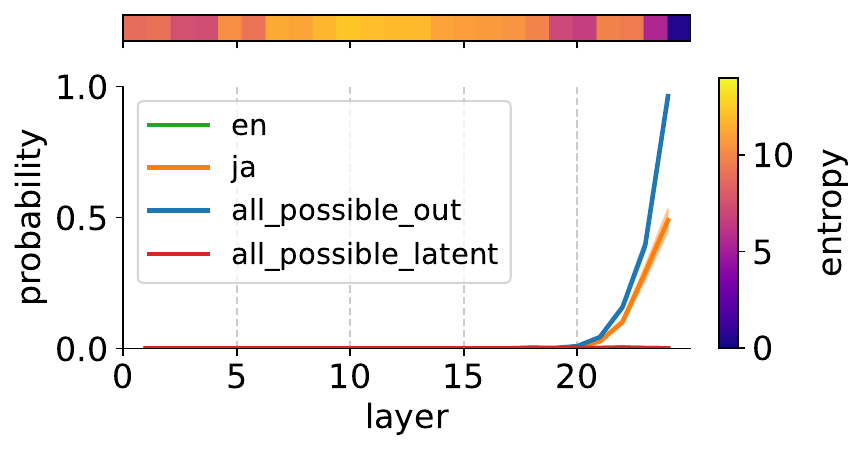}
    \subcaption{Japanese After}
    \label{fig:qwen1.8b-ja-after}
    \end{minipage}
\caption{\textbf{Logit lens on Qwen1.5-1.8B in English, Chinese, and Japanese scenarios (languages not in training data).} The horizontal axes is the layer num and the vertical axes is the probability. "en" (Green) means the latent English output corresponding to the correct answer in the target language. "en/zh/ja" (Orange) means the correct answer in the target language. "all\_possible\_out" (Blue) means the probability of all possible outputs in the target language. "all\_possible\_latent" (Red) means all possible outputs in English.}
\label{fig:qwen1.8b-logit-lens}
\end{figure*}

We report the logit lens and the PCA results of Qwen1.5-1.8B (total 24 layers) here. In Figure \ref{fig:qwen1.8b-logit-lens}, we can find that as we mentioned above, while utilizing logit lens on Qwen1.5, a non-English-centric model, there is no intermediate latent output before generating the output in the target language finally. This indicates that logit lens might not be an effective tool for analyzing the non-English-centric LLMs.

\begin{figure*}[tbp]
    \centering
    \begin{minipage}[b]{0.24\textwidth}
    \includegraphics[width=1.0\linewidth]{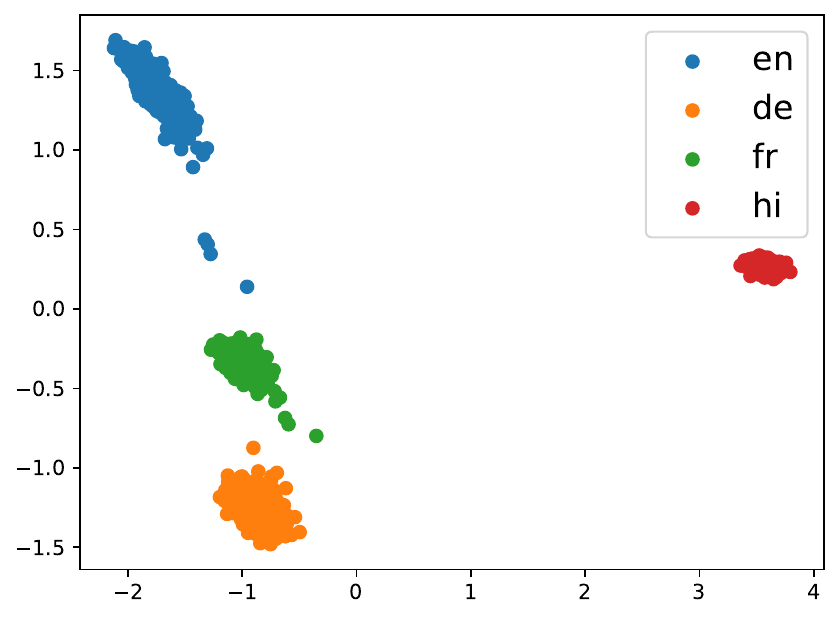}
    \subcaption{Layer 6 Before}
    \label{fig:qwen1.8b-pca-6-before}
    \end{minipage}
    \begin{minipage}[b]{0.24\textwidth}
    \includegraphics[width=1.0\linewidth]{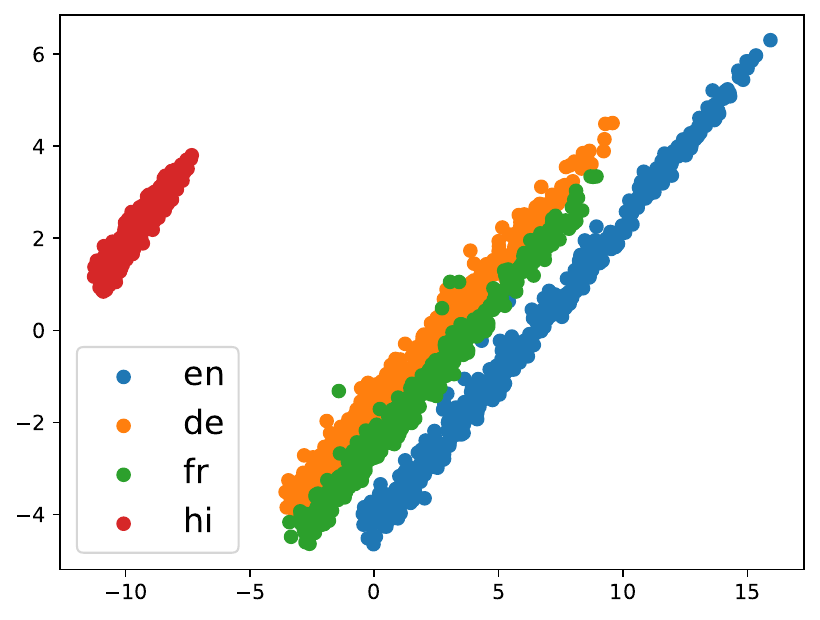}
    \subcaption{Layer 12 Before}
    \label{fig:qwen1.8b-pca-12-before}
    \end{minipage}
    \begin{minipage}[b]{0.24\textwidth}
    \includegraphics[width=1.0\linewidth]{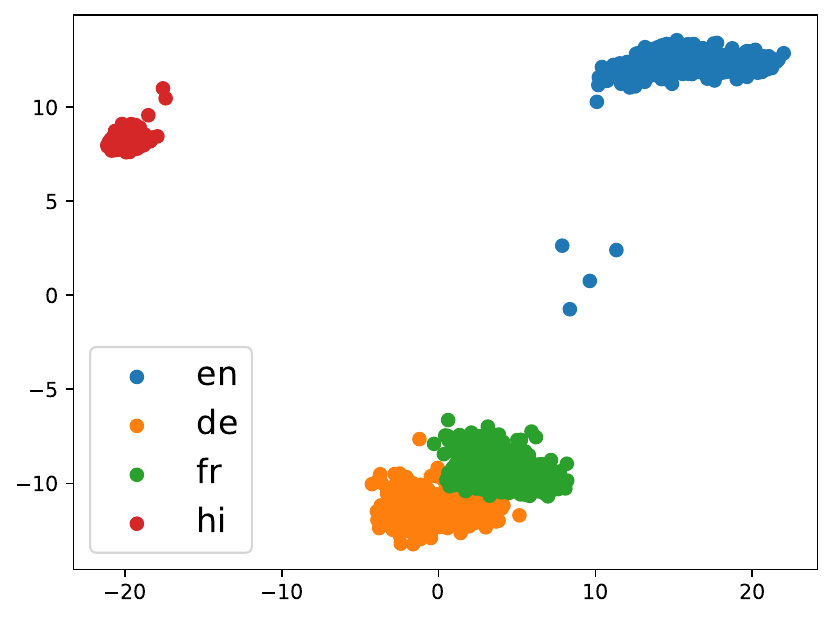}
    \subcaption{Layer 18 Before}
    \label{fig:qwen1.8b-pca-18-before}
    \end{minipage}
    \begin{minipage}[b]{0.24\textwidth}
    \includegraphics[width=1.0\linewidth]{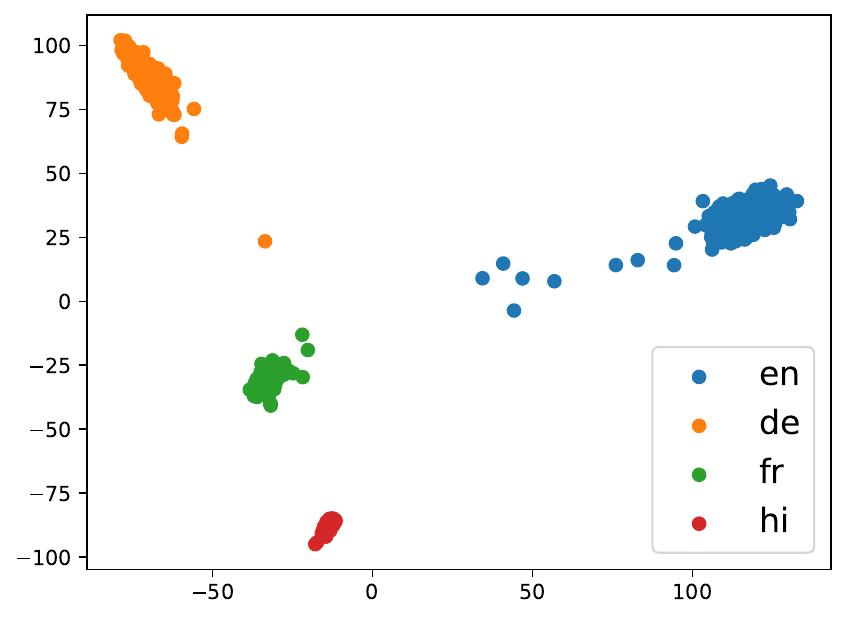}
    \subcaption{Layer 24 Before}
    \label{fig:qwen1.8b-pca-24-before}
    \end{minipage}
    \begin{minipage}[b]{0.24\textwidth}
    \includegraphics[width=1.0\linewidth]{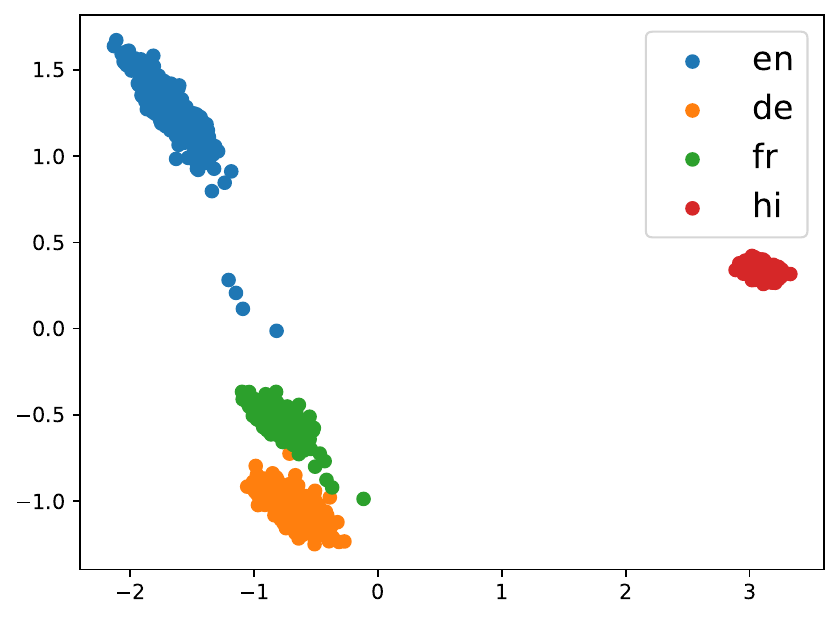}
    \subcaption{Layer 6 After}
    \label{fig:qwen1.8b-pca-6-after}
    \end{minipage}\begin{minipage}[b]{0.24\textwidth}
    \includegraphics[width=1.0\linewidth]{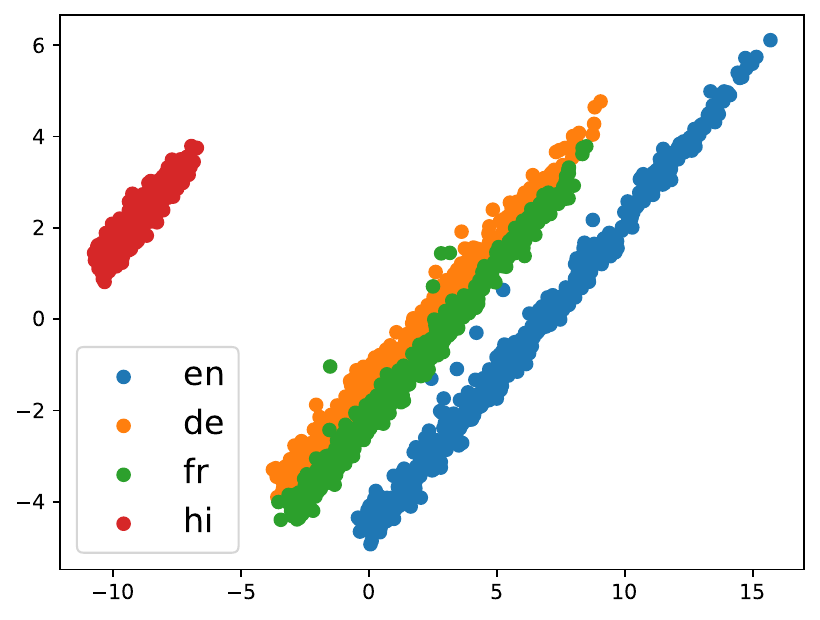}
    \subcaption{Layer 12 After}
    \label{fig:qwen1.8b-pca-12-after}
    \end{minipage}\begin{minipage}[b]{0.24\textwidth}
    \includegraphics[width=1.0\linewidth]{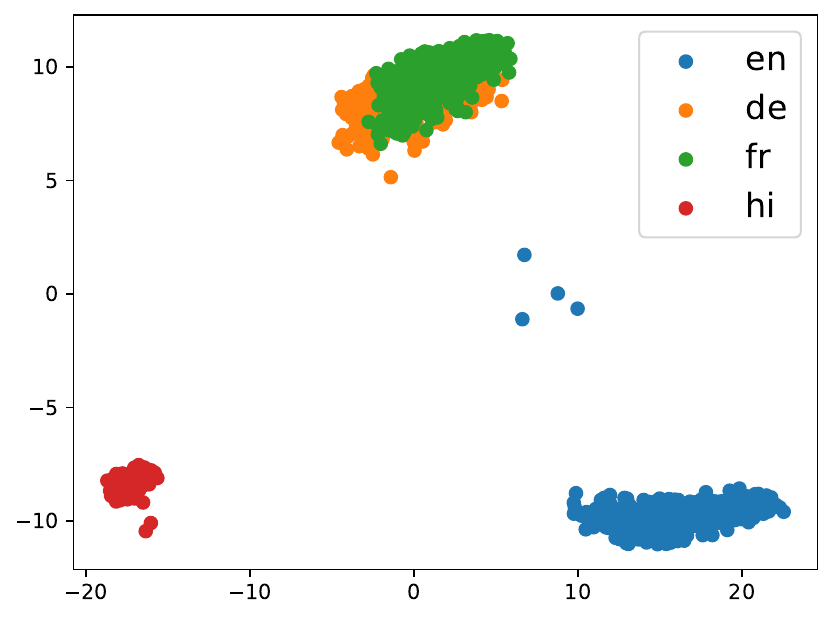}
    \subcaption{Layer 18 After}
    \label{fig:qwen1.8b-pca-18-after}
    \end{minipage}\begin{minipage}[b]{0.24\textwidth}
    \includegraphics[width=1.0\linewidth]{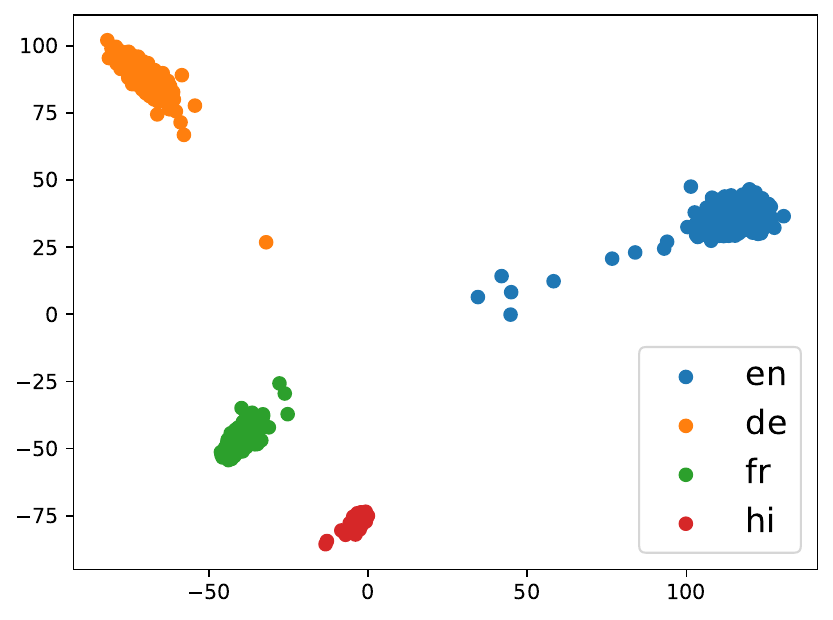}
    \subcaption{Layer 24 After}
    \label{fig:qwen1.8b-pca-24-after}
    \end{minipage}
\caption{\textbf{PCA (Principal Component Analysis) on Qwen1.5-1.8B in English, German, French and Hindi scenarios.} Before means the base model. After means the trained model. All logits are mapped into the two-dimensional representation. Each point in the plot corresponds to one instance. \textbf{There is also a similar latent representation pattern for different languages in the intermediate layers while logit lens can't reveal it.}}
\label{fig:qwen1.8b-pca}
\end{figure*}

We further report the PCA results in Figure \ref{fig:qwen1.8b-pca}, which also indicates a clear similar latent representation pattern for different languages in the non-English-centric LLMs' intermediate layers. This further reinforces the significance of multilingual alignment, which also provides the basis for the success of question alignment paradigm on Qwen. The Pearson coefficient results reported in Table \ref{tab:pearson-qwen1.8b-12} and Table \ref{tab:pearson-qwen1.8b-18} show the better alignment with English, consisting with the results of Mistral-7B.

\begin{table}[htbp]
    \centering
    \begin{tabular}{l|cc}
        \toprule
        \textbf{Layer 12} & \textbf{Base} & \textbf{Trained} \\
        \midrule
        en-de & 0.9816 & 0.9820 \\
        en-fr & 0.9848 & 0.9850 \\
        en-hi & 0.9730 & 0.9742 \\
        de-fr & 0.9852 & 0.9855 \\
        de-hi & 0.9760 & 0.9769 \\
        fr-hi & 0.9732 & 0.9740 \\
        en-th & 0.9538 & 0.9547 \\
        en-sw & 0.9746 & 0.9763 \\
        en-ms & 0.9815 & 0.9823 \\
        \bottomrule
    \end{tabular}
    \caption{Pearson correlation coefficient of 1 dimension PCA results in Qwen1.5-1.8B layer 12.}
    \label{tab:pearson-qwen1.8b-12}
\end{table}

\begin{table}[htbp]
    \centering
    \begin{tabular}{l|cc}
        \toprule
        \textbf{Layer 18} & \textbf{Base} & \textbf{Trained} \\
        \midrule
        en-de & 0.7331 & 0.9176 \\
        en-fr & 0.6243 & 0.9303 \\
        en-hi & 0.3623 & 0.8589 \\
        de-fr & 0.7493 & 0.9477 \\
        de-hi & 0.5639 & 0.8965 \\
        fr-hi & 0.4729 & 0.8854 \\
        en-th & 0.2881 & 0.4566 \\
        en-sw & -0.0424 & 0.8592 \\
        en-ms & 0.4297 & 0.8648 \\
        \bottomrule
    \end{tabular}
    \caption{Pearson correlation coefficient of 1 dimension PCA results in Qwen1.5-1.8B layer 18.}
    \label{tab:pearson-qwen1.8b-18}
\end{table}

\section{Full Results of SNLI and PAWS}\label{appendix:full-results-snli-paws}
We report the complete results of SNLI and PAWS on 20 different languages in Table \ref{tab:nli-all} and \ref{tab:paws-all} separately. Similar to the emotion classification task, we can see that models instruction-tuned on multilingual translation data significantly outperform the base model, which confirms that our findings have good generalization across different tasks.

\begin{table*}[tbp]
    \centering
    \begin{tabular}{l|cccccccccc}
        \toprule
        \textbf{Qwen1.5-14B} & \textbf{en} & \textbf{zh} & \textbf{de} & \textbf{fr} & \textbf{es} & \textbf{it} & \textbf{nl} & \textbf{ja} & \textbf{ru} & \textbf{sv} \\
        \midrule
        base & 84.50 & 83.50 & 74.17 & 75.17 & 81.17 & 78.67 & 78.17 & 51.17 & 76.83 & 76.17 \\
        zh/es $\Rightarrow$ en & \textbf{92.50} & 84.67 & 82.67 & 82.83 & \textbf{85.50} & \textbf{83.83} & \textbf{84.67} & 57.00 & \textbf{82.67} & \textbf{84.33} \\
        zh/de $\Rightarrow$ en & 91.83 & 84.50 & \textbf{83.67} & \textbf{84.50} & 85.00 & 83.67 & 84.50 & \textbf{57.17} & 81.67 & \textbf{84.33} \\
        zh/it $\Rightarrow$ en & 91.67 & 83.83 & 80.83 & 82.67 & 84.00 & 80.00 & 83.50 & 55.83 & 81.50 & 83.33 \\
        sw/hi $\Rightarrow$ en & 91.33 & \textbf{85.67} & 80.67 & 80.83 & 83.50 & 81.50 & 82.33 & 55.00 & 79.50 & 82.83 \\
        \bottomrule
        \toprule
        \textbf{Qwen1.5-14B} & \textbf{sl} & \textbf{pl} & \textbf{bg} & \textbf{no} & \textbf{ms} & \textbf{is} & \textbf{hi} & \textbf{th} & \textbf{sw} & \textbf{bn} \\
        \midrule
        base & 63.17 & 67.83 & 64.33 & \textbf{43.00} & 75.00 & 48.17 & 61.00 & 69.67 & 45.83 & 41.33 \\
        zh/es $\Rightarrow$ en & 66.00 & 76.83 & 76.33 & 37.50 & \textbf{80.67} & \textbf{57.67} & 71.33 & \textbf{75.00} & \textbf{58.33} & 40.33 \\
        zh/de $\Rightarrow$ en & \textbf{66.83} & 77.00 & \textbf{78.00} & 35.33 & 80.50 & 57.50 & \textbf{73.00} & \textbf{75.00} & \textbf{58.33} & \textbf{43.33} \\
        zh/it $\Rightarrow$ en & 65.67 & \textbf{77.33} & 76.17 & 36.67 & 79.00 & 56.00 & 70.50 & 73.67 & 55.33 & 41.33 \\
        sw/hi $\Rightarrow$ en & 63.00 & 76.67 & 72.17 & 39.67 & 80.33 & 54.17 & 67.67 & 74.67 & 56.83 & 41.33 \\
        \bottomrule
    \end{tabular}
    \caption{Accuracy of Qwen1.5-14B base model and trained models on the SNLI. We report all of the results on 20 languages. The accuracy of randomly choosing is 33.33\%. We highlight the best results for every language.}
    \label{tab:nli-all}
\end{table*}

\begin{table*}[tbp]
    \centering
    \begin{tabular}{l|cccccccccc}
        \toprule
        \textbf{Qwen1.5-14B} & \textbf{en} & \textbf{zh} & \textbf{de} & \textbf{fr} & \textbf{es} & \textbf{it} & \textbf{nl} & \textbf{ja} & \textbf{ru} & \textbf{sv} \\
        \midrule
        base & 85.4 & 78.4 & 74.0 & 82.2 & 78.8 & 79.4 & 78.6 & 75.0 & 73.0 & 76.6 \\
        zh/es $\Rightarrow$ en & \textbf{87.8} & \textbf{83.2} & 81.2 & \textbf{86.0} & \textbf{85.8} & 83.8 & \textbf{86.0} & \textbf{77.0} & 82.6 & 84.4 \\
        zh/de $\Rightarrow$ en & 87.0 & 80.8 & \textbf{84.8} & 84.8 & 84.6 & 82.2 & \textbf{86.0} & \textbf{77.0} & \textbf{83.0} & 84.6 \\
        zh/it $\Rightarrow$ en & 87.4 & 80.8 & 80.4 & 85.0 & 83.8 & \textbf{85.2} & 85.2 & \textbf{77.0} & 82.2 & \textbf{85.2} \\
        sw/hi $\Rightarrow$ en & 87.2 & 78.4 & 79.6 & 81.8 & 84.6 & 83.2 & 84.4 & 76.6 & 81.0 & 85.0 \\
        \bottomrule
        \toprule
        \textbf{Qwen1.5-14B} & \textbf{sl} & \textbf{pl} & \textbf{bg} & \textbf{no} & \textbf{ms} & \textbf{is} & \textbf{hi} & \textbf{th} & \textbf{sw} & \textbf{bn} \\
        \midrule
        base & 70.2 & 79.8 & 75.2 & 78.6 & 82.0 & 67.0 & 73.2 & 79.4 & 75.4 & 65.0 \\
        zh/es $\Rightarrow$ en & \textbf{82.6} & 80.8 & 82.0 & 86.4 & \textbf{86.8} & 77.4 & 80.6 & \textbf{82.0} & 79.2 & 76.2 \\
        zh/de $\Rightarrow$ en & 80.8 & 80.2 & 82.0 & 85.8 & 85.8 & 77.0 & \textbf{81.4} & 81.4 & 80.8 & 74.2 \\
        zh/it $\Rightarrow$ en & 81.2 & 81.6 & 81.8 & 86.8 & 85.6 & 77.2 & 79.4 & 81.2 & \textbf{81.2} & 74.8 \\
        sw/hi $\Rightarrow$ en & 82.4 & \textbf{81.8} & \textbf{82.6} & \textbf{87.0} & 85.8 & \textbf{78.2} & 80.8 & 80.4 & 78.2 & \textbf{77.0} \\
        \bottomrule
    \end{tabular}
    \caption{Accuracy of Qwen1.5-14B base model and trained models on the PAWS. We report all of the results on 20 languages. The accuracy of randomly choosing is 50.0\%. We highlight the best results for every language.}
    \label{tab:paws-all}
\end{table*}

\end{document}

%% file: Chapters/00-abstract.tex
\begin{abstract}
Recently, Large Language Models (LLMs) have shown impressive language capabilities. While most of the existing LLMs have very unbalanced performance across different languages, multilingual alignment based on translation parallel data is an effective method to enhance the LLMs' multilingual capabilities. In this work, we 
discover and comprehensively investigate the spontaneous multilingual alignment improvement of LLMs. We find that LLMs instruction-tuned on the question translation data (i.e. without annotated answers) are able to encourage the alignment between English and a wide range of languages, even including those unseen during instruction-tuning. Additionally, we utilize different settings and mechanistic interpretability methods to analyze the LLM's performance in the multilingual scenario comprehensively. Our work suggests that LLMs have enormous potential for improving multilingual alignment efficiently with great language and task generalization.
\footnote{Our code and data is available at: \url{https://github.com/Shimao-Zhang/LLM-Multilingual-Learner}.}
\end{abstract}

%% file: Chapters/01-introduction.tex
\section{Introduction}
Large Language Models (LLMs) have recently shown impressive language capabilities across numerous downstream language tasks~\citep{zhao2023survey}. However, most existing LLMs are trained on extensive high-resource languages text~\citep{touvron2023llama, brown2020language, jiang2023mistral}, which lead to a significant performance gap between high-resource languages and low-resource languages~\citep{huang2023not, zhang2023plug, gao2024multilingual}. For the same task and question contents, using different languages for inputs may have a significant impact on the model's performance.

Some studies have conducted comprehensive exploration about how to enhance the LLMs' capabilities across different languages. The classical approach typically follows the translate-based paradigm~\citep{liu2024translation}.
Considering LLMs' great performance on the high-resource languages, some cross-lingual alignment and transfer methods are proposed~\citep{eronen2023zero, zhu2024question, zhao2024llama}. Question alignment~\citep{zhu2024question} is an outstanding paradigm among these methods which effectively improves multilingual alignment at lower cost, i.e. only utilizes the X-English parallel question translation data.

Meanwhile, some studies have further explored the LLMs, revealing that English also participate in the intermediate latent reasoning of these models even when LLMs are prompted in non-English~\citep{wendler2024llamas, zhao2024large}. These findings suggest that for LLMs, different languages are not isolated, and LLMs are able to leverage the connections between various languages to address problems in the multilingual scenarios. Researchers also reveal the shared semantic space for different languages~\citep{chang2022geometry}, which is consistent with the findings above and indicates the importance of the multilingual alignment. Also, \citet{kew2023turning} discover that multilingual instruction-tuning with three languages improves model’s cross-lingual transfer abilities on some generative tasks.

Intuitively, LLMs have abilities to acclimatize themselves to the multilingual environment through appropriate training~\citep{shi2022language}.
Many existing methods rely on instruction-tuning on the multilingual instruction-tuning datasets~\citep{kew2023turning, liu2024translation}. However, given the question alignment paradigm, 
utilizing multilingual alignment is also helpful for improving LLMs' multilingual abilities. Additionally, we focus on question alignment in our work to eliminate the interference of task-related data with annotated answers from our analysis of multilingual alignment. Based on the findings above, can LLMs achieve better multilingual alignment across different languages efficiently through appropriate methods?

In this work, we investigate the multilingual alignment of LLMs, where we only train the LLMs on the parallel data without annotated answers (only queries) in a few languages. Following question alignment, we conduct the experiments on models in different types (English-centric or not) and parameter sizes, and test across a wide range of languages on different benchmarks. We find that question alignment following~\citet{zhu2024question} can effectively enhance the multilingual capabilities of LLMs, which indicates that models can effectively utilize the relevant knowledge and capabilities learned during the pretraining process with question alignment, consisting with the "Superficial Alignment Hypothesis"~\citep{zhou2024lima}. Our results also indicate that conducting question alignment in a small number of languages brings significantly better multilingual alignment even between English and many languages unseen during instruction-tuning process, which implies good language generalization. Furthermore, we also use logit lens~\citep{logit-lens} and dimensionality reduction techniques~\citep{pearson1901liii} to study the latent states of LLMs, providing more comprehensive perspectives and empirical results for the alignment improvements in our experiments.

%% file: Chapters/02-background.tex
\section{Background}

\subsection{Unbalanced Multilingual Performance}\label{subsec:unbalanced-performance}
With a much larger number of parameters pretrained on a massive corpus, LLMs have shown the impressive capabilities in a variety of language tasks~\citep{zhao2023survey}. These models are mainly pretrained on English data, which often accounts for 90\% or even more of all training data.
We present a partial language distribution of LLaMA-2's training data in Table \ref{tab:LLaMA2-lang-percentage} in Appendix \ref{appendix:llama2-lang-distribution}. Meanwhile, most of the LLMs also show unstable and unbalanced performance in multilingual scenarios, especially for some low-resource languages~\citep{zhang2023don, zhu2024question}. It's important to enable LLMs to adapt to a wider range of users and scenarios.

\subsection{Cross-lingual Enhancement for Large Language Models}\label{subsec:x-lingual transfer}
While LLMs still exhibit significant shortcomings in multilingual scenarios, many researchers propose multilingual LLMs that are specifically adjusted for multilingual tasks~\citep{team2023internlm, le2023bloom, wei2023polylm}. But for multilingual LLMs, researches indicate a decline in their performance in English because of the limited tokens and parameter size~\citep{lin2022few, scao2022language}.

Based on the existing LLMs, researchers have made great efforts to enhancing the multilingual performance, which include two categories: prompting close-source LLMs and instruction-tuning open-source LLMs. For the former, some studies utilize translation-based strategies which translate the non-English input into English firstly before solving the problem~\citep{huang2023not, qin2023cross}. This type of approaches are constrained by the translation quality of the model itself and is cumbersome for users.

For the latter, LLMs shows significant improvement of multilingual abilities and good task generalization through multilingual multitask fine-tuning~\citep{muennighoff2022crosslingual}. \citet{chen2023breaking} follow the translation-based approach and instruction-tune the model on a multilingual version of GSM8K, which is translated from English GSM8K~\citep{cobbe2021training}. \citet{liang2024machine} create a new intermediate language MUL (Machine-created Universal Language) as a translatable unified representation of shared concepts across different languages. "X-English" parallel question translation data have also been used for multilingual question alignment~\citep{zhu2024question}. In our work, we mainly analyse based on the question alignment, which is an outstanding alignment methods, and eliminates the interference of the annotated answers from our analysis.

\subsection{Mechanistic Interpretability}\label{subsec: mechanistic interpretability}
In addition to improving the performance of LLMs, it is also crucial to understand and explain the principles of neural networks and related methods explicitly. Current works mainly analyze LLMs' actions by observing the internal states during the inference process. Intermediate logits and neuron activation states are both important objects of observation.

Although the English-centric LLMs are mainly trained on English data, they also show good performance across some non-English languages~\citep{shi2022language}. Logit lens~\citep{logit-lens} is an early proposed technique that using the model head in the final layer to project the intermediate latent logits directly to the vocabulary space. It have been evidenced that LLaMA 2~\citep{touvron2023llama}, a open-source English-centric LLMs, have explicit English output in its latent states even when having non-English inputs~\citep{wendler2024llamas}. There is also a hypothesis about how LLMs handle multilingualism that LLMs will solve task by English with the help of multilingual knowledge, and output in the target language finally~\citep{zhao2024large}. All these results indicate that there are connections between various languages for LLMs, and LLMs have the capability to spontaneously learn to utilize multiple languages to solve problems. \citet{zhao2024large} calculate the overlapping ratio of the language-specific neurons of different languages in different layers. The results indicate that neurons belonging to different languages exhibit clear distribution differences. In our experiments, we utilize logit lens and dimensionality reduction techniques to help us better understand the mechanism behind our findings.

%% file: Chapters/03-methods.tex
\section{Analysis Pipeline}\label{sec:analysis-pipeline}
We investigate the effect of question translation parallel data on LLMs' performance across a wide range of languages even unseen during the fine-tuning process.

We define the universal set of languages as $\mathbf{U}$:
\begin{equation}
    \mathbf{U} = \{l_0,\ l_1,\ l_2,\ ...\ ,\ l_{n-1}\}
\end{equation}
where $l_i$ is the $i$-th language in $\mathbf{U}$, $n$ is the total number of languages. We let $l_0$ refer to English specially here.

We select a few of non-English languages $\mathcal{L}_s = \{l_i, ... , l_k\} \subseteq \mathbf{U}$, and a target language $l_t \in \mathbf{U}$, $l_t \notin \mathcal{L}_s$. Then we will construct translation parallel data from every language $l \in \mathcal{L}_s$ to target language $l_t$. When construct the translation data, we only use the questions without annotated answers. Then we get a translation dataset $\mathcal{Q}_{train}$ including source question $\mathcal{Q}_s$ and the corresponding target question $\mathcal{Q}_t$, which means $\mathcal{Q}_{train} = \{(q_s, q_t) \mid q_s \in \mathcal{Q}_s\ and\ q_t \in \mathcal{Q}_t\}$. We instruct-tune the model on the translation task:
\begin{equation}
    \mathop{\arg\min}_{\theta}\sum_{(q_s, q_t) \in \mathcal{Q}_{train}}-\operatorname{log}p_{\theta}(q_t \mid q_s)
\end{equation}
where $\theta$ is the model parameters, $\mathcal{Q}_{train}$ is the whole training translation dataset, $q_s$ is the question in the source language, $q_t$ is the question in the target language. Then we get the trained model:
\begin{equation}
    \theta^{'} = \theta + \Delta \theta
\end{equation}

We use question translation data for training to eliminate the impact of annotated answers themselves. And we use in-context learning for test while the model haven't been trained on the corresponding task.

We test the trained model on all languages $l \in \mathbf{U}$. We construct the testing dataset $\mathcal{Q}_{test} = \{\mathcal{Q}_l \mid l \in \mathbf{U}\}$ for every language, where $\mathcal{Q}_l$ consists of all test questions in the language $l$.
\begin{equation}
    \mathrm{Accuracy}_l = \sum_{q \in \mathcal{Q}_l}\mathbf{I}_{\theta^{'}}(\hat{a} = a \mid q)
\end{equation}
\begin{equation}
    \mathrm{Accuracy} = \frac{\sum_{l \in \mathbf{U}} \mathrm{Accuracy}_l}{\vert \mathbf{U} \vert}
\end{equation}
where $\mathbf{I}$ is a function that takes 1 when the proposition is true and 0 otherwise. $\mathcal{Q}_l$ denotes the test dataset of language $l$. $\mathbf{U}$ is the universal set of languages we use in our work. $\hat{a}$ is the answer that the model predicts base on $q$, and $a$ is the golden answer corresponding to $q$.

%% file: Chapters/04-experimental-setup.tex
\section{Experimental Setup}\label{sec:experimental-setup}
We conduct our experiments on both English-centric and non-English-centric models. And we utilize different representative tasks and different model parameter sizes to further strengthen our conclusions. In this section, we introduce our experimental settings in detailed.

\begin{table*}[tbp]
    \centering
    \small
    \begin{tabular}{l|cccccccccc}
        \toprule
        \textbf{Mistral-7B} & \textbf{en} & \textbf{zh} & \textbf{de} & \textbf{fr} & \textbf{es} & \textbf{it} & \textbf{nl} & \textbf{ja} & \textbf{ru} & \textbf{sv} \\
        \midrule
        base & 89.2 & 92.4 & 91.8 & 93.4 & 94.2 & 93.8 & 93.6 & 93.0 & 93.2 & 93.4 \\
        zh $\Rightarrow$ en & 95.2 & 94.8 & 94.8 & \textbf{95.2} & 94.4 & 94.4 & 94.8 & 94.4 & 94.0 & \textbf{95.4} \\
        sw $\Rightarrow$ en & \textbf{95.4} & 93.4 & 94.2 & 94.4 & 94.2 & 94.4 & 93.0 & 93.6 & 93.8 & 94.8 \\
        zh/es $\Rightarrow$ en & 95.2 & 95.0 & \textbf{95.0} & 95.0 & 94.8 & 92.8 & 94.6 & \textbf{95.0} & \textbf{94.4} & 94.8 \\
        zh/de $\Rightarrow$ en & 95.2 & 95.4 & 94.8 & \textbf{95.2} & \textbf{95.2} & \textbf{95.2} & 94.8 & 93.6 & 94.2 & 94.6 \\
        zh/it $\Rightarrow$ en & \textbf{95.4} & \textbf{95.8} & 94.8 & 94.0 & \textbf{95.2} & 92.6 & 94.4 & 93.0 & 94.2 & 95.2 \\
        sw/hi $\Rightarrow$ en & \textbf{95.4} & 94.6 & 94.4 & 93.4 & 93.4 & 93.6 & 93.6 & 94.0 & 93.8 & 94.4 \\
        sw/th $\Rightarrow$ en & \textbf{95.4} & 95.0 & 93.8 & 93.4 & 93.4 & 92.8 & 93.6 & 92.6 & 93.2 & 94.0 \\
        zh/es/ru $\Rightarrow$ en & \textbf{95.4} & 95.4 & 94.4 & 94.0 & 94.6 & 92.6 & 94.6 & 94.2 & 94.0 & 94.2 \\
        zh/de/it $\Rightarrow$ en & 95.2 & 95.6 & 94.4 & 95.0 & 94.0 & 93.8 & \textbf{95.0} & 93.6 & 94.2 & 94.6 \\
        \bottomrule
        \toprule
        \textbf{Mistral-7B} & \textbf{sl} & \textbf{pl} & \textbf{bg} & \textbf{no} & \textbf{ms} & \textbf{is} & \textbf{hi} & \textbf{th} & \textbf{sw} & \textbf{bn} \\
        \midrule
        base & 87.6 & 93.2 & 91.6 & 92.4 & 91.8 & 63.2 & 81.6 & 83.0 & 58.0 & 71.0 \\
        zh $\Rightarrow$ en & \textbf{94.0} & 94.0 & \textbf{94.6} & 92.2 & 89.0 & 84.0 & 88.8 & \textbf{88.4} & 75.8 & \textbf{81.0} \\
        sw $\Rightarrow$ en & 89.8 & 92.6 & 93.6 & 93.4 & 90.0 & 72.0 & 64.4 & 51.4 & \textbf{81.2} & 54.0 \\
        zh/es $\Rightarrow$ en & 93.2 & 93.6 & 94.0 & 93.0 & 92.2 & 81.2 & 87.0 & 84.8 & 75.6 & 75.4 \\
        zh/de $\Rightarrow$ en & 93.4 & 94.0 & \textbf{94.6} & \textbf{93.6} & 92.2 & \textbf{86.6} & 84.8 & \textbf{88.4} & 71.8 & 68.6 \\
        zh/it $\Rightarrow$ en & 92.6 & 93.8 & 94.2 & \textbf{93.6} & \textbf{92.6} & 84.2 & 77.6 & 77.2 & 71.6 & 60.0 \\
        sw/hi $\Rightarrow$ en & 89.2 & 93.0 & 93.2 & 92.6 & 90.0 & 71.8 & \textbf{89.8} & 87.0 & 77.6 & 79.4 \\
        sw/th $\Rightarrow$ en & 92.8 & 92.0 & 93.2 & 87.2 & 84.4 & 79.4 & 86.8 & 84.0 & 81.0 & 74.2 \\
        zh/es/ru $\Rightarrow$ en & 93.6 & \textbf{94.2} & 93.4 & 93.4 & 91.4 & 83.8 & 85.0 & 86.0 & 77.0 & 76.0 \\
        zh/de/it $\Rightarrow$ en & 91.2 & 93.6 & 94.2 & 93.4 & 91.8 & 83.2 & 77.2 & 82.4 & 69.0 & 71.4 \\
        \bottomrule
    \end{tabular}
    \caption{Accuracy of Mistral-7B base model and aligned models on the Amazon Reviews Polarity. We report at least two sets of results for each language quantity to strengthen our conclusions. The accuracy of randomly choosing is 50.0\%. "X/Y/Z $\Rightarrow$ T" means using a randomly mixed dataset including 10k X to T, 10k Y to T, 10k Z to T translation data for instruction-tuning. We highlight the best results for every language.}
    \label{tab:mistral-emo}
\end{table*}

\paragraph{Models}
We choose representative open-source LLMs for our experiments:
\begin{itemize}
    \item \textbf{Mistral:} Mistral-7B-v0.1~\citep{jiang2023mistral} is an advanced open-source English-centric large language model, which is one of the most popular open-source LLMs.
    \item \textbf{Qwen:} To enhance the generalization and reliability of our conclusions, we also choose models of different types and parameter sizes. Qwen1.5 is a newly released and enhanced version of Qwen~\citep{bai2023qwen}. Qwen is pretrained on a multilingual dataset with a significant portion of the data being in English and Chinese, which means it is not an English-centric model. We choose Qwen1.5-1.8B, Qwen1.5-4B, Qwen1.5-14B for our experiments.
\end{itemize}

\paragraph{Datasets}
Following \citet{wendler2024llamas}, we select test tasks based on two fundamental principles: 
\begin{enumerate}
    \item \textbf{Obvious Answers:} Obvious answers reduce the entropy during the inference process, minimizing the impact of irrelevant tokens on our analysis.
    \item \textbf{Fixed Answers:} Fixed answers (as opposed to open-ended responses) provide clearer observation targets, facilitating analysis through observing the latent outputs of the model. Deterministic outputs also make it easier for us to control the model's outputs.
\end{enumerate}

Based on these, we conduct our experiments on three different types of tasks:
\begin{itemize}
    \item \textbf{Emotion Classification: }Emotion classification is an important and classic NLP task~\citep{alswaidan2020survey}, which always has three common outputs: "positive", "negative", and "neutral". We choose Amazon Reviews Polarity\footnote{\url{https://huggingface.co/datasets/amazon_polarity}}~\citep{zhang2015character}, a famous dataset includes two classes "positive" and "negative", to construct the parallel data mentioned in \S \ref{subsec:x-lingual transfer} and the test data. We extract 10K instances from train subset for parallel data and 500 instances from test subset for test data respectively.
    \item \textbf{Natural Language Inference: }Natural language inference (NLI) aims to judge the relationship between a given premise and a hypothesis sentence. There are always three possible outputs: "entailment", "neutral", and "contradiction". We choose SNLI\footnote{\url{https://huggingface.co/datasets/stanfordnlp/snli}} (Stanford Natural Language Inference)~\citep{bowman2015large} to conduct our experiments. Following the emotion classification task, we extract 10K instances from train subset for parallel data and 600 instances from test subset for test data respectively.
    \item \textbf{Paraphrase Identification: }Model needs to judge if two given sentences are semantically equivalent in the paraphrase identification task, which includes two possible labels. We conduct our experiments on PAWS\footnote{\url{https://huggingface.co/datasets/google-research-datasets/paws}} dataset~\citep{zhang2019paws}, which is a famous dataset proposed by Google. Following the above tasks, we extract 10K instances from train subset for parallel data and 500 instances from test subset for test data respectively.
\end{itemize}

\paragraph{Languages}
We conduct our following experiments across 20 languages in this work. As shown in Table \ref{tab:LLaMA2-lang-percentage} in Appendix \ref{appendix:llama2-lang-distribution}, we choose English (en), German (de), French (fr), Swedish (sv), Chinese (zh), Spanish (es), Russian (ru), Dutch (nl), Italian (it), and Japanese (ja) as the top 10 highest-resource languages according to \citet{touvron2023llama}. Additionally, we choose another 10 representative languages to strengthen our work, including Slovenian (sl), Polish (pl), Bulgarian (bg), Norwegian (no), Malay (ms), Icelandic (is), Hindi (hi), Thai (th), Swahili (sw), and Bengali (bn).

\paragraph{Implementations}
We all use LoRA~\citep{hu2021lora} to instruction-tune the pre-trained models on the mixed parallel translation data first. We train LLMs on the translation data excluding the golden answers to mitigate the impact of the data of the tasks themselves on the model's capabilities. We use in-context learning which not only doesn't interfere with LLMs' parameters but also help LLMs handle the tasks better. We use constrained decoding rather than sampling that is used for diverse generation~\citep{zhang2024edt} to eliminate the interference of irrelevant outputs on the results.
More details are shown in Appendix \ref{appendix:experimental-implementations}.

%% file: Chapters/05-results.tex
\begin{table*}[tbp]
    \centering
    \small
    \begin{tabular}{l|cccc}
        \toprule
        \textbf{Model} & \textbf{Qwen1.5-1.8B} & \textbf{Qwen1.5-4B} & \textbf{Mistral-7B} & \textbf{Qwen1.5-14B} \\
        \midrule
        base & 68.35 & 79.52 & 87.07 & 86.27 \\
        zh/es $\Rightarrow$ en & \textbf{76.13} & 81.99 & \textbf{90.83} & 91.53 \\
        zh/de $\Rightarrow$ en & 74.23 & 82.64 & 90.81 & \textbf{92.25} \\
        zh/it $\Rightarrow$ en & 75.70 & 83.32 & 89.10 & 92.13 \\
        sw/hi $\Rightarrow$ en & 75.37 & \textbf{85.32} & 90.21 & 90.28 \\
        \bottomrule
    \end{tabular}
    \caption{Average accuracy of models of different parameter sizes on the Amazon Reviews Polarity. We highlight the best results for every model.}
    \label{tab:emo-all}
\end{table*}

\begin{table*}[tbp]
    \centering
    \small
    \begin{tabular}{l|ccc}
        \toprule
        \textbf{Qwen1.5-14B} & \textbf{Emotion Classification} & \textbf{NLI} & \textbf{Paraphrase Identification} \\
        \midrule
        base & 86.27 & 66.94 & 76.36 \\
        zh/es $\Rightarrow$ en & 91.53 & 73.03 & \textbf{82.59} \\
        zh/de $\Rightarrow$ en & \textbf{92.25} & \textbf{73.28} & 82.21 \\
        zh/it $\Rightarrow$ en & 92.13 & 71.94 & 82.15 \\
        sw/hi $\Rightarrow$ en & 90.28 & 71.48 & 81.80 \\
        \bottomrule
    \end{tabular}
    \caption{Average accuracy results of Qwen1.5-14B base model and trained models on the Amazon Reviews Polarity, SNLI and PAWS across 20 different languages. The accuracy of randomly choosing is 33.33\% for SNLI and 50.00\% for the other two tasks. We highlight the best results for every task. Full results are reported in Appendix \ref{appendix:full-results-snli-paws}.}
    \label{tab:nli-paws-avg}
\end{table*}

\section{Results}\label{sec:results}
In this section, we report the main results across different experimental settings and conduct some discussions based on the results.

\begin{table}[tbp]
    \centering
    \small
    \begin{tabular}{l|cc}
        \toprule
        \textbf{Model} & \textbf{Qwen1.5-1.8B} & \textbf{Mistral-7B} \\
        \midrule
        base & 68.35 & 87.07 \\
        ja/it $\Rightarrow$ zh & 73.32 & 90.34 \\
        sw/hi $\Rightarrow$ zh & \textbf{77.00} & \textbf{90.58} \\
        en/ja $\Rightarrow$ zh & 71.92 & 90.04 \\
        \bottomrule
    \end{tabular}
    \caption{Average accuracy on Amazon Reviews Polarity. We replace English with Chinese as the target language. We highlight the best results for each model.}
    \label{tab:non-english-alignment-emo}
\end{table}

\subsection{Main Results}
We report the accuracy of Mistral-7B on emotion classification task in Table \ref{tab:mistral-emo}. Clearly, we can see that the models trained on multilingual translation data outperform the original model significantly across a lot of languages, which indicates that model have much stronger multilingual capabilities after a multilingual training. We summarize our empirical findings as follows:

\begin{enumerate}
    \item \textbf{Large language models can learn to handle multilingualism better spontaneously.} Traditionally, fine-tuning or alignment on the target languages is needed to help the model adapt. However, our results indicate that LLMs are able to perform effective learning and transfer across multiple languages without parallel data for most of them. As seen, models has much higher overall accuracy across 20 languages after training on data containing 2-4 languages.

    \item \textbf{High-resource languages are not only good learners but also good leaders.} Is there any difference when we use high-resource languages or low-resource languages in our training data? Our results in Table \ref{tab:mistral-emo} show that 
    the accuracy on high-resource language is not significantly related to whether the corresponding language data is used. More importantly, training on high-resource language data enables the model to achieve more stable improvements across multiple languages compared to that on low-resource languages (Swahili, Hindi, and Thai).

    \item \textbf{A few languages can lead to spontaneous multilingual learning.} We select one, two, three languages with English for instruction-tuning respectively. In Table \ref{tab:mistral-emo}, although using more languages sometimes leads to more stable improvements, model trained only on Chinese and English have achieved similar overall performance improvements. This is also consistent with the findings of~\citet{kew2023turning}. The multilingual alignment improvement shows great language generalization.

    \item \textbf{Our findings remain consistent across models of different parameter sizes.} We also present the average accuracy results of Qwen1.5-1.8B, Qwen1.5-4B, and Qwen1.5-14B in Table \ref{tab:emo-all} to strengthen our conclusions. We find significant multilingual performance improvements across all of these models.
\end{enumerate}

We have also validated our findings on the other two tasks to strengthen our conclusions, including Natural Language Inference (NLI) and Paraphrase Identification. The model needs to determine the relationship between two paragraphs of text in both of these two tasks. We conduct our experiment on SNLI for NLI task and PAWS for Paraphrase Identification task. We report the average accuracy of Qwen1.5-14B across all languages in Table \ref{tab:nli-paws-avg}. And we report the full results on each language in Appendix \ref{appendix:full-results-snli-paws}.

\subsection{Analysis}
Building upon the above results, we conduct more comprehensive observations and analyses of the model's behavior.

\paragraph{English is not necessary as the target language in the training data.}
As elaborated in Section \ref{sec:experimental-setup}, we use outputs in English uniformly for all languages in our previous experiments. English has been widely used for multilingual transfer as a pivot language~\citep{zhu2024question, hu2023large}. We further investigate the case of replacing English with Chinese in training data and report the results in Table \ref{tab:non-english-alignment-emo}. Mistral and Qwen1.5 represent two different types of LLMs (English-centric or not) respectively. From the results, we can find that using Chinese as the target language leads to the same conclusions as using English. For both of the two types of LLMs, using Chinese rather than English as the target language is also helpful for models' multilingual performance improvement, which indicates that English is not necessary as the target language in the training data.

\begin{figure*}[tbp]
    \centering
    \begin{minipage}[b]{0.32\textwidth}
    \includegraphics[width=1.0\linewidth]{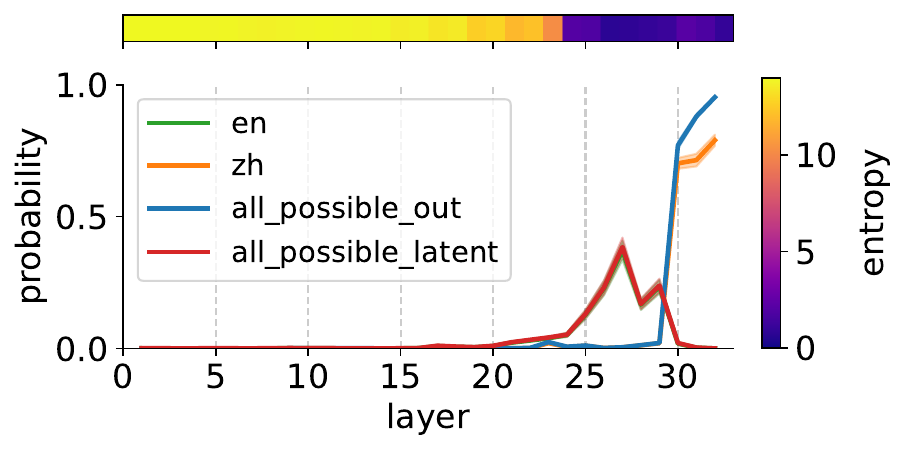}
    \subcaption{Chinese Before}
    \label{fig:zh-before}
    \end{minipage}
    \begin{minipage}[b]{0.32\textwidth}
    \includegraphics[width=1.0\linewidth]{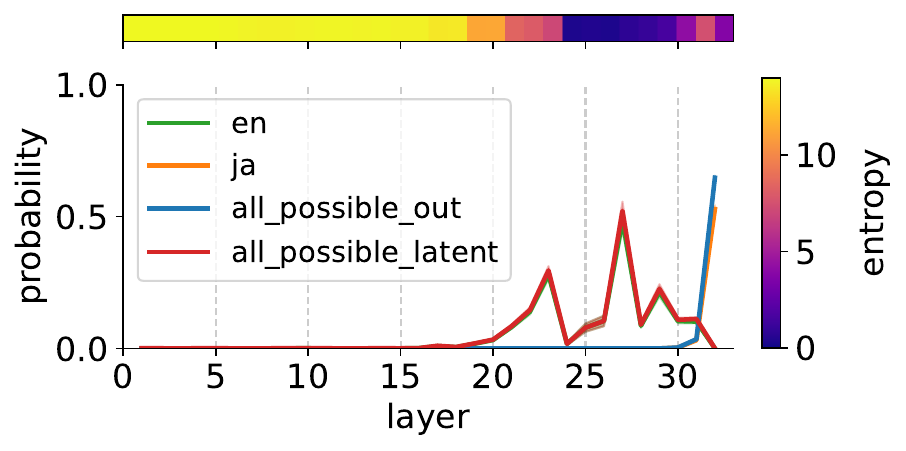}
    \subcaption{Japanese Before}
    \label{fig:ja-before}
    \end{minipage}
    \begin{minipage}[b]{0.32\textwidth}
    \includegraphics[width=1.0\linewidth]{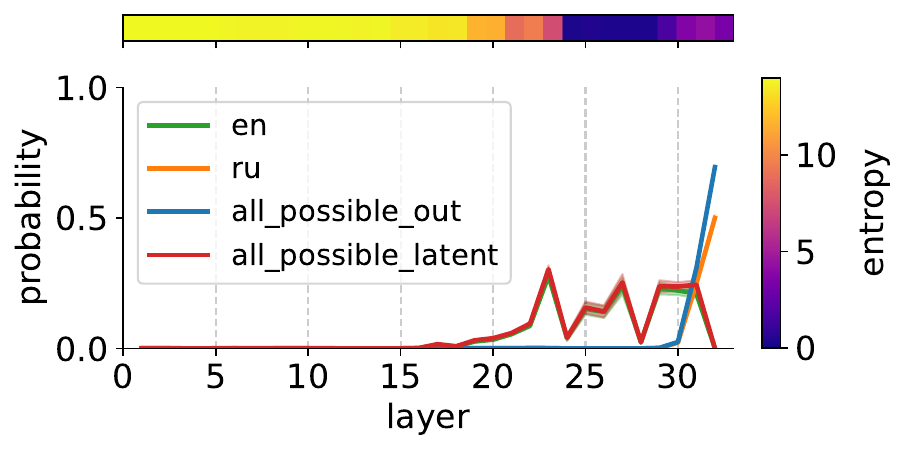}
    \subcaption{Russian Before}
    \label{fig:ru-before}
    \end{minipage}
    \begin{minipage}[b]{0.32\textwidth}
    \includegraphics[width=1.0\linewidth]{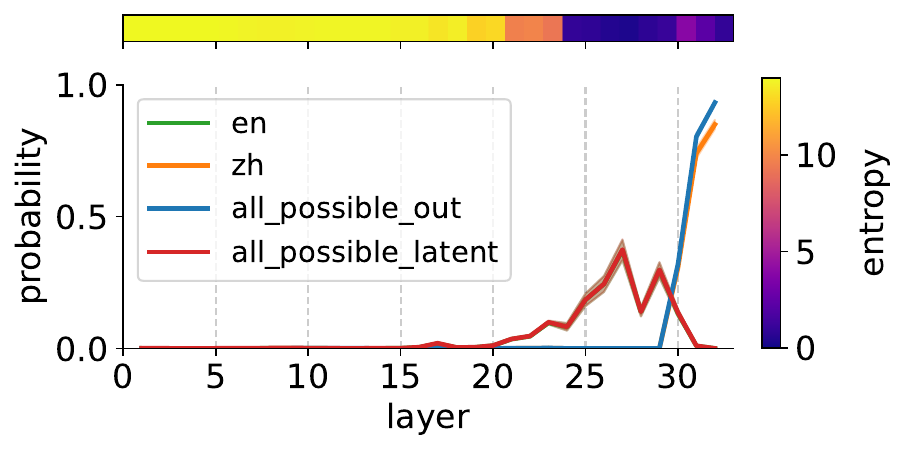}
    \subcaption{Chinese After}
    \label{fig:zh-after}
    \end{minipage}
    \begin{minipage}[b]{0.32\textwidth}
    \includegraphics[width=1.0\linewidth]{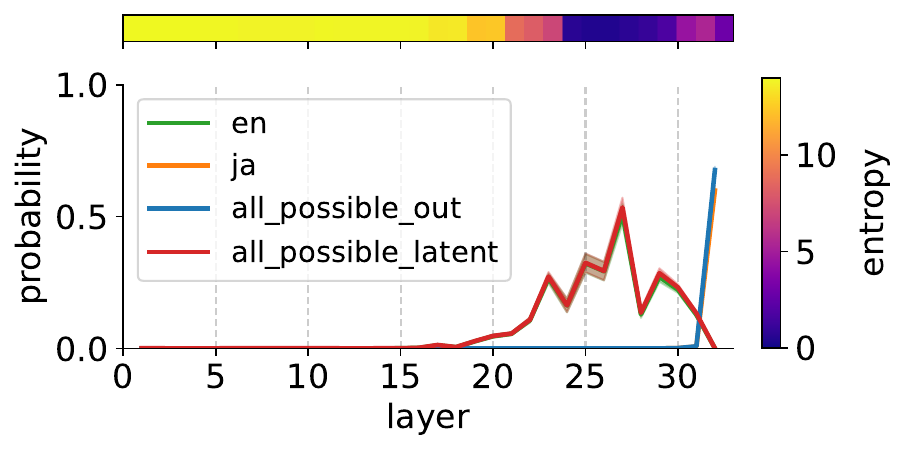}
    \subcaption{Japanese After}
    \label{fig:ja-after}
    \end{minipage}
    \begin{minipage}[b]{0.32\textwidth}
    \includegraphics[width=1.0\linewidth]{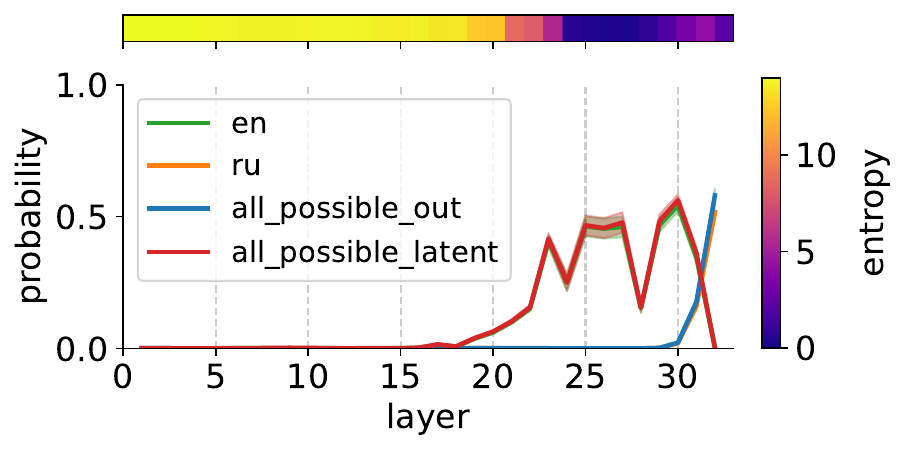}
    \subcaption{Russian After}
    \label{fig:ru-after}
    \end{minipage}
\caption{\textbf{Logit lens on Mistral-7B in Chinese, Japanese and Russian scenarios (languages not in training data).} The horizontal axes is the layer num and the vertical axes is the probability. "en" (Green covered by Red) means the latent English output corresponding to the correct answer in the target language. "zh/ja/ru" (Orange) means the correct answer in the target language. "all\_possible\_out" (Blue) means the probability of all possible outputs in the target language. "all\_possible\_latent" (Red) means all possible outputs in English.}
\label{fig:logit-lens}
\end{figure*}

\paragraph{It is not necessary but more beneficial to use the train subset corresponding to the test data as the source of translation data.}
Following \citet{zhu2024question}, in our previous experiments, we construct the parallel translation data for instruction-tuning based on the train subset corresponding to the test dataset, which have the similar data characteristics and distributions. We further cross-test the Qwen1.5-14B trained on SNLI  on Amazon Reviews Polarity and the Qwen1.5-14B trained on Amazon Reviews Polarity on SNLI. We report the results in Table \ref{tab:cross-text-different-tasks}.
We can find that although the models trained on data with different distributions also have better overall performance in most cases, they have a worse performance than that trained on the data corresponding to the target task. That means the multilingual data is crucial for enhancing the model's multilingual capabilities, and similar types of data is more helpful. This is consistent with the "Superficial Alignment Hypothesis"~\citep{zhou2024lima}, which indicates that model learns knowledge and capabilities almost entirely in pretraining process, while alignment only guides the model to utilize the different "subdistribution of formats". So the data in the same subdistribution of formats is more beneficial.

\begin{table}[tbp]
    \centering
    \small
    \begin{tabular}{l|cc}
        \toprule
        \textbf{Model} & \textbf{Amazon Polarity} & \textbf{SNLI} \\
        \midrule
        base & 86.27 & 66.94 \\
        zh/es $\Rightarrow$ en & 90.38 & 68.72 \\
        zh/de $\Rightarrow$ en & 90.75 & 67.50 \\
        zh/it $\Rightarrow$ en & 90.46 & 67.76 \\
        sw/hi $\Rightarrow$ en & 90.53 & 65.76 \\
        \bottomrule
    \end{tabular}
    \caption{The model tested on Amazon Reviews Polarity is trained on SNLI questions. The model tested on SNLI is trained on Amazon Reviews Polarity questions.}
    \label{tab:cross-text-different-tasks}
\end{table}

\begin{table}[tbp]
    \centering
    \small
    \begin{tabular}{l|cc}
        \toprule
        \textbf{Model} & \textbf{Same Language} & \textbf{Task-agnostic} \\
        \midrule
        base & 76.86 & 50.40 \\
        zh/es $\Rightarrow$ en & 83.48 & \textbf{77.61}  \\
        zh/de $\Rightarrow$ en & 83.69 & 72.28  \\
        zh/it $\Rightarrow$ en & 82.33 & 72.32  \\
        sw/hi $\Rightarrow$ en & \textbf{84.59} & 74.92  \\
        \bottomrule
    \end{tabular}
    \caption{The results of Mistral-7B on emotion classification task for different output types. \textbf{Same Language} means the outputs in the same language with the inputs. \textbf{Task-agnostic} means using the task-agnostic outputs.}
    \label{tab:different-types-outputs}
\end{table}

\paragraph{How about using outputs in different types?}
Except the outputs in English, we also conduct our experiments by using outputs in different types, including outputs in the same language with the inputs and task-agnostic outputs. When using outputs in the same language with the inputs, as shown in Table \ref{tab:different-types-outputs}, the model also perform better after instruction-tuning, while performing worse compared to using English outputs (shown in Table \ref{tab:emo-all}) under the same settings. This confirms our conclusion in Section \ref{sec:experimental-setup} that generating content in the target language is sometimes another
great challenge for LLMs except understanding and solving multilingual problems themselves.

We further replace "positive" with "ox" and replace "negative" with "horse" to investigate the cases of using task-agnostic outputs. We report the results in Table \ref{tab:different-types-outputs}. Firstly, we can observe a significant decrease in multilingual performance of the base model when using task-agnostic outputs, which indicates that task-specific outputs are important for effective in-context learning (ICL). Clearly, we find a significant improvement in multilingual performance of the instruction-tuned models. By comparing the results before and after training, we can find that our training greatly improves the model's ICL capability on the specific task, and this capability improvement exhibits excellent multilingual generalization. Based on the Superficial Alignment Hypothesis, we infer that the questions in only a few languages are able to effectively activate the corresponding subdistribution of formats across a wide range of languages.

\begin{figure*}[tbp]
    \centering
    \begin{minipage}[b]{0.24\textwidth}
    \includegraphics[width=1.0\linewidth]{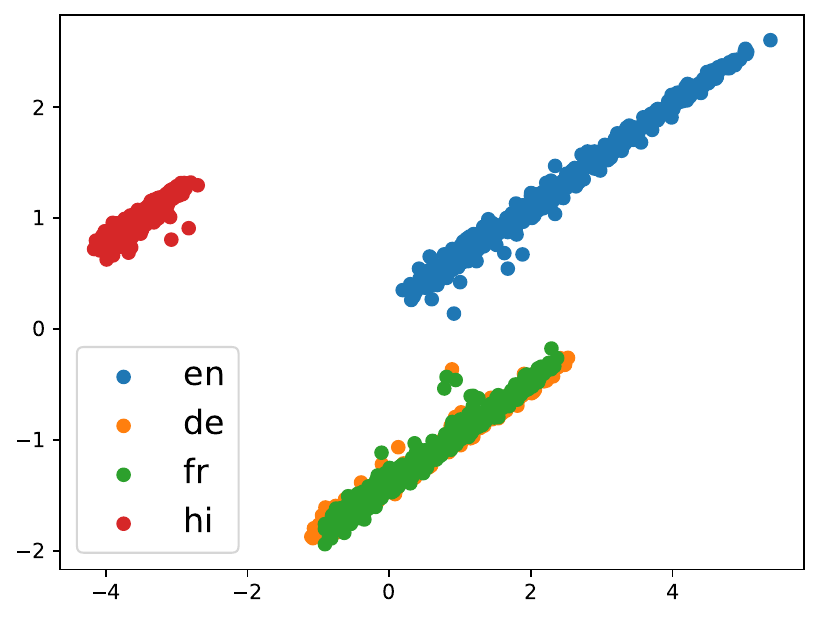}
    \subcaption{Layer 20 Before}
    \label{fig:pca-20-before}
    \end{minipage}
    \begin{minipage}[b]{0.24\textwidth}
    \includegraphics[width=1.0\linewidth]{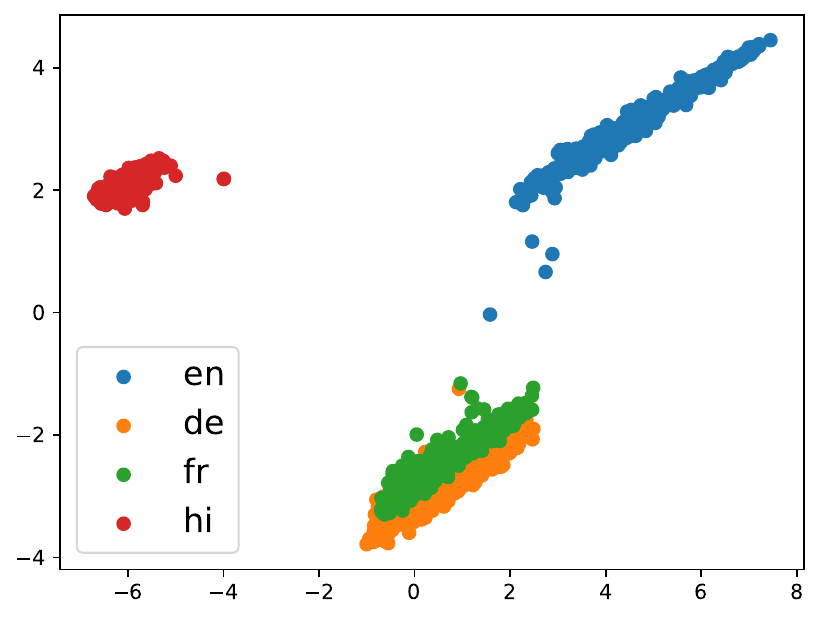}
    \subcaption{Layer 25 Before}
    \label{fig:pca-25-before}
    \end{minipage}
    \begin{minipage}[b]{0.24\textwidth}
    \includegraphics[width=1.0\linewidth]{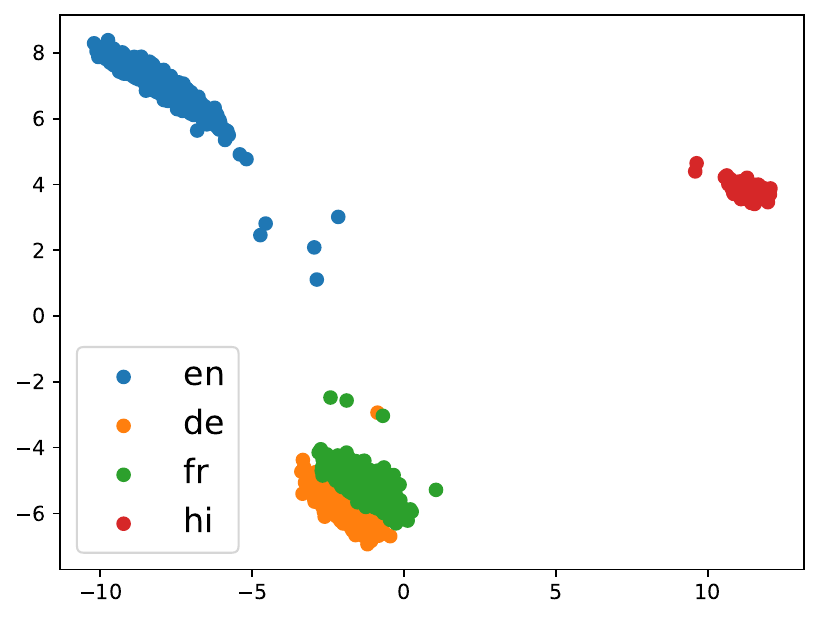}
    \subcaption{Layer 30 Before}
    \label{fig:pca-30-before}
    \end{minipage}
    \begin{minipage}[b]{0.24\textwidth}
    \includegraphics[width=1.0\linewidth]{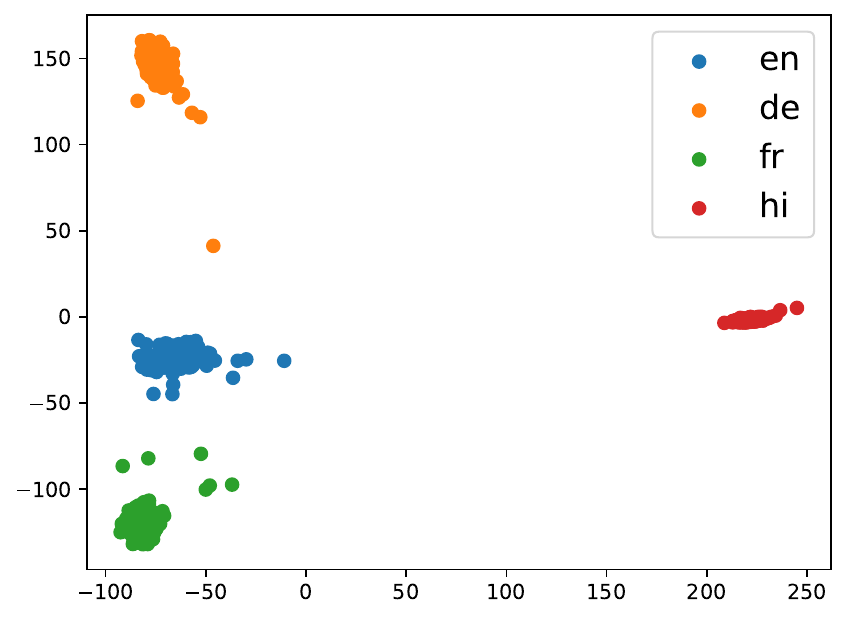}
    \subcaption{Layer 32 Before}
    \label{fig:pca-32-before}
    \end{minipage}
    \begin{minipage}[b]{0.24\textwidth}
    \includegraphics[width=1.0\linewidth]{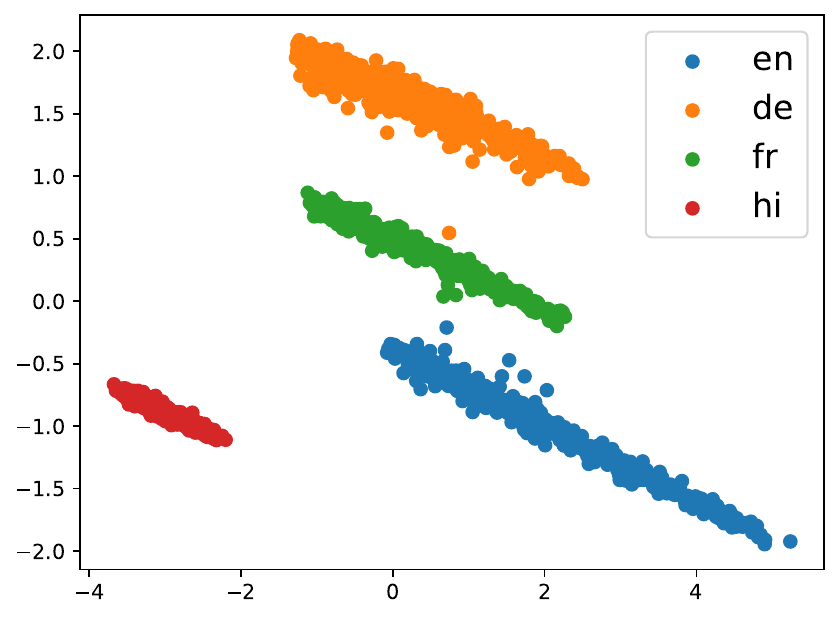}
    \subcaption{Layer 20 After}
    \label{fig:pca-20-after}
    \end{minipage}\begin{minipage}[b]{0.24\textwidth}
    \includegraphics[width=1.0\linewidth]{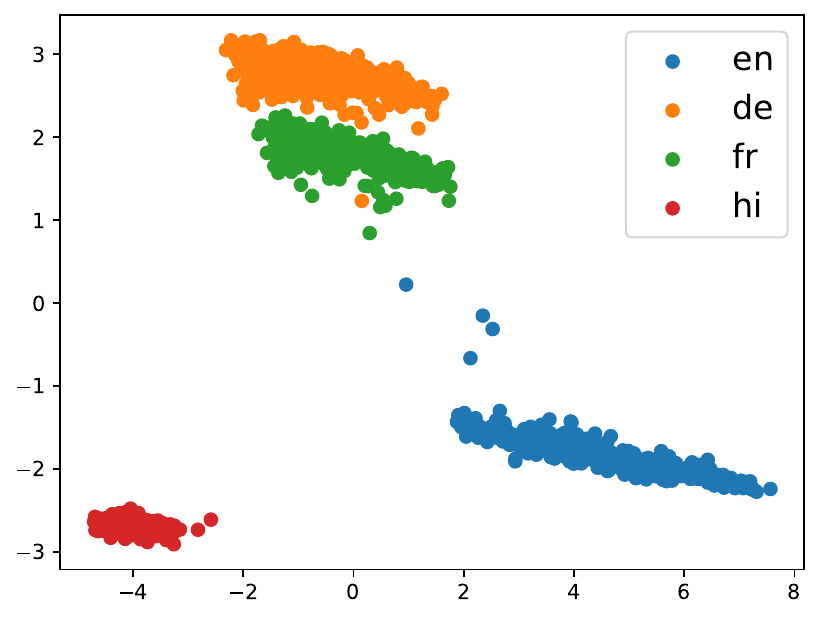}
    \subcaption{Layer 25 After}
    \label{fig:pca-25-after}
    \end{minipage}\begin{minipage}[b]{0.24\textwidth}
    \includegraphics[width=1.0\linewidth]{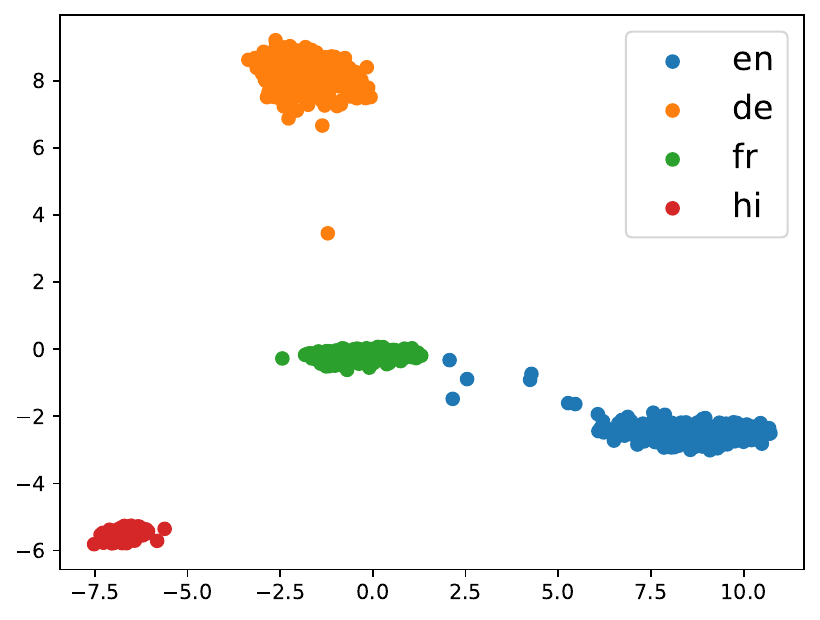}
    \subcaption{Layer 30 After}
    \label{fig:pca-30-after}
    \end{minipage}\begin{minipage}[b]{0.24\textwidth}
    \includegraphics[width=1.0\linewidth]{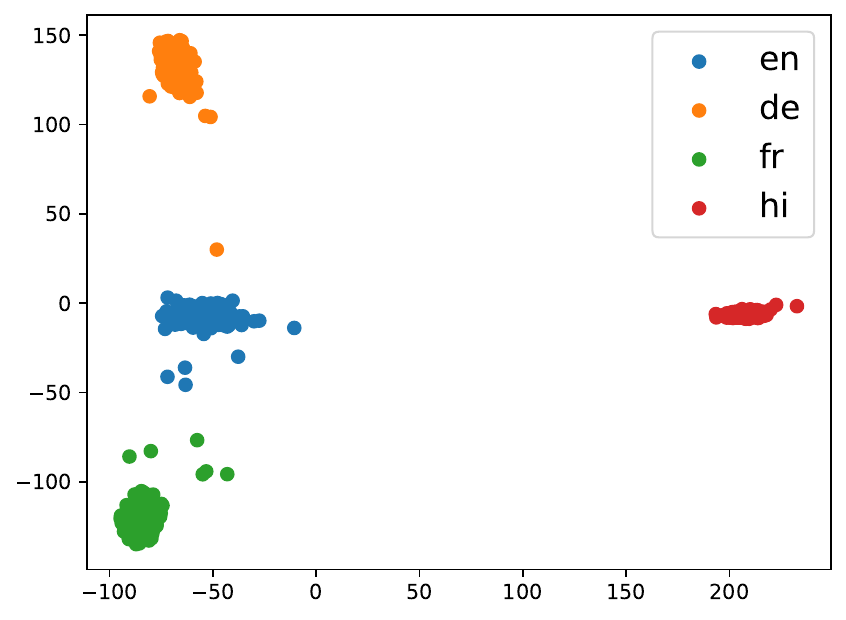}
    \subcaption{Layer 32 After}
    \label{fig:pca-32-after}
    \end{minipage}
\caption{\textbf{PCA (Principal Component Analysis) on Mistral-7B in English, German, French and Hindi scenarios.} Before means the base model. After means the trained model. All logits are mapped into the two-dimensional representation. Each point in the plot corresponds to one instance.}
\label{fig:pca}
\end{figure*}

%% file: Chapters/06-further-analysis.tex
\section{Mechanistic Interpretability Analysis}\label{sec:mechanistic interpretability-analysis}
In this section, we further utilize methods mentioned in \S \ref{subsec: mechanistic interpretability} to analyze the model's changes before and after the training.

\subsection{Logit Lens}\label{subsec:logit-lens-analysis}
Following \citet{wendler2024llamas}, we utilize \textit{logit lens} to analyze the changes of the model.
We utilize logit lens on Qwen1.5, a series of LLMs that are not English-centric, and find there is not English latent outputs in the intermediate layers. And the prefix token overlapping between target language and English will also bring errors to the results. So we choose Chinese, Japanese and Russian as three representative languages for our experiment, which shows significant improvement in our results before. Following \citet{wendler2024llamas}, we use the outputs in the same language with the inputs (Table \ref{tab:different-types-outputs}). We conduct our experiments on Mistral-7B and its best trained version "sw/hi $\Rightarrow$ en" in Table \ref{tab:different-types-outputs}. We report the results in Figure \ref{fig:logit-lens}. Clearly, we can observe the following points: (1) All models generate latent English output before generating outputs in the target language finally; (2) The proportion of the probability of the correct answer increases in the sum of all possible answer probabilities; (3) The probability of all other possible answers (except correct answer) in the latent English outputs is nearly zero; (4) The area of latent English output significantly increases, which means the trained models perform latent inference in English better and indicates better alignment.

\subsection{Principal Component Analysis}\label{subsec:PCA-analysis}
We further utilize the dimensionality reduction technique to visualize the intermediate layer latent outputs of the model across different languages. PCA (Principal Component Analysis)~\citep{pearson1901liii} can be used in some scenarios where logit lens doesn't work. Principal components are a few linear combinations of the original variables that maximally explain the variance of all the variables~\citep{greenacre2022principal}. We utilize PCA to map the latent logits into the two-dimensional representation. Based on the patterns shown in Figure \ref{fig:logit-lens}, we report layer 20, 25, 30 and the last layer as four representative layers in Figure \ref{fig:pca}. We have the following findings: (1) The points of different languages follow the similar patterns in layer 20 and layer 25, where English latent outputs have appeared and outputs in the target language haven't appeared. We further calculate the Pearson correlation coefficient of 1 dimension PCA results (Appendix \ref{appendix:pearson-pca}). There is a strong linear correlation between representations of different languages, which also indicates an uniform latent representation pattern during inference process; (2) Representations belong to different languages exhibit greater distance from each other after training; (3) The results of the last layer is similar because of the same possible outputs; (4) Based on Pearson coefficient reported in Appendix \ref{appendix:pearson-pca}, the correlation between low-resource languages (hi, th, sw, ms) and other high-resource languages significantly improves, which suggests better alignment with English.

%% file: Chapters/07-conclusion.tex
\section{Conclusion}\label{sec:conclusion}
In this paper, we find that LLMs only trained on question translation data without annotated answers are able to get a significant multilingual alignment improvement between English and a wide range of languages, even those unseen during instruction-tuning. We conduct the experiments on different models, different benchmarks and 20 different languages to strengthen our conclusions. Our results indicate that utilizing question alignment significantly enhances the multilingual alignment and the in-context learning capabilities of LLMs. And these improvements demonstrate the excellent model and language generalization. Furthermore, we also conduct comprehensive analysis based on some mechanistic interpretability methods, including logit lens and dimensionality reduction technique. Our work demonstrates the enormous potential of LLMs for efficient multilingual capability improvement. We hope our work can inspire the community to further explore this promising direction for the better multilingual alignment.

\section{Limitations}\label{sec:limitations}
We aim to draw more attention to the multilingual alignment which is a promising research direction. Despite our work has demonstrated LLMs' strong capability of multilingual generalization and the great potential of efficient multilingual alignment, there are still some limitations waiting for research. Because we investigate the models trained on the parallel question translation data in our work to eliminate the interference of the task-related data with annotated answers from our analysis of alignment, we utilize few-shot learning to help model handle the target tasks better. Analyzing LLMs' multilingual alignment in a zero-shot setting properly would further strengthen the conclusions if possible.

Due to the limited resources, we conduct experiments on different LLM scale from 1.8B to 14B in this work. We are willing to verify our conclusions on larger LLMs (70B or larger) if more resources are available in the future. Meanwhile, we mainly utilize automatic translation engine in our work because of the limited resources, while data annotated by native speakers would be more accurate.
